\documentclass[conference]{IEEEtran}
\IEEEoverridecommandlockouts
\usepackage{multirow}
\usepackage{soul}
\usepackage{mathtools}
\usepackage{amsmath,amsfonts,amssymb,amsthm}
\usepackage{algorithmic}
\usepackage[ruled,vlined]{algorithm2e}
\usepackage{graphicx}
\usepackage{float}
\usepackage{lipsum}
\usepackage{subfig}
\usepackage{textcomp}
\usepackage{xcolor}
\usepackage{placeins}
\usepackage{hhline}
\usepackage{tabularx}
\usepackage{enumitem}
\usepackage{setspace}
\usepackage{booktabs}
\usepackage{adjustbox}
\usepackage[normalem]{ulem}
\usepackage{indentfirst}
\usepackage{titlesec}
\usepackage{color, colortbl}

\makeatletter
\newcommand{\mathleft}{\@fleqntrue\@mathmargin0pt}
\newcommand{\mathcenter}{\@fleqnfalse}

\makeatother

\newtheorem{definition}{\textbf{Definition}}

\def\BibTeX{{\rm B\kern-.05em{\sc i\kern-.025em b}\kern-.08em
    T\kern-.1667em\lower.7ex\hbox{E}\kern-.125emX}}

\definecolor{Gray}{gray}{0.85}
\definecolor{LightCyan}{rgb}{0.88,1,1}

\newcolumntype{a}{>{\columncolor{Gray}}c}
\newcolumntype{b}{>{\columncolor{white}}c}

\begin{document}
\title{DeepGAR: Deep Graph Learning for Analogical Reasoning}

\author{\IEEEauthorblockN{Chen Ling\IEEEauthorrefmark{1}\IEEEauthorrefmark{5},
Tanmoy Chowdhury\IEEEauthorrefmark{2}\IEEEauthorrefmark{5}, Junji Jiang\IEEEauthorrefmark{3}, Junxiang Wang\IEEEauthorrefmark{1},\\ Xuchao Zhang\IEEEauthorrefmark{4}, Haifeng Chen\IEEEauthorrefmark{4}, and Liang Zhao\IEEEauthorrefmark{1}\IEEEauthorrefmark{6}}
\IEEEauthorblockA{\IEEEauthorrefmark{1} Emory University\quad \IEEEauthorrefmark{2} George Mason University \quad \IEEEauthorrefmark{3} Tianjin University \quad \IEEEauthorrefmark{4} NEC Labs America}
\IEEEauthorblockA{Email: \IEEEauthorrefmark{1}\{chen.ling, junxiang.wang, liang.zhao\}@emory.edu,
\IEEEauthorrefmark{2}tchowdh6@gmu.edu,\\
\IEEEauthorrefmark{3}anjou\_j@tju.edu.cn,
\IEEEauthorrefmark{4}\{xuczhang, haifeng\}@nec-labs.com}
\IEEEauthorblockA{\IEEEauthorrefmark{5} Equal Contribution\quad \quad \IEEEauthorrefmark{6} Corresponding Author}
}

\maketitle

\begin{abstract}
    Analogical reasoning is the process of discovering and mapping correspondences from a target subject to a base subject. As the most well-known computational method of analogical reasoning, Structure-Mapping Theory (SMT) abstracts both target and base subjects into relational graphs and forms the cognitive process of analogical reasoning by finding a corresponding subgraph (i.e., correspondence) in the target graph that is aligned with the base graph. However, incorporating deep learning for SMT is still under-explored due to several obstacles: 1) the combinatorial complexity of searching for the correspondence in the target graph; 2) the correspondence mining is restricted by various cognitive theory-driven constraints. To address both challenges, we propose a novel framework for Analogical Reasoning (DeepGAR) that identifies the correspondence between source and target domains by assuring cognitive theory-driven constraints. Specifically, we design a geometric constraint embedding space to induce subgraph relation from node embeddings for efficient subgraph search. Furthermore, we develop novel learning and optimization strategies that could end-to-end identify correspondences that are strictly consistent with constraints driven by the cognitive theory. Extensive experiments are conducted on synthetic and real-world datasets to demonstrate the effectiveness of the proposed DeepGAR over existing methods. The code and data are available at: https://github.com/triplej0079/DeepGAR.

    


\end{abstract}


\section{Introduction}
    Analogical reasoning is a cognitive process of transferring information from a particular subject (the analog or source) to another (the target). Over the last two decades, analogical reasoning has gained prominence in artificial intelligence research for both practical and theoretical reasons \cite{hall1989computational}. In solving and learning new problems, analogical reasoning is capable of overcoming the immense search complexity of finding solutions to novel problems or inducing generalized knowledge from experience \cite{vamvakoussi2019use, lu2019emergence}. Furthermore, analogical reasoning may provide theoretical justification for the shift from low-level visual processing to abstract conceptual change in domains like natural language understanding and visual perception \cite{sagi2012difference, gentner1997analogical}.

    Interests in analogical reasoning have spawned extensive computational approaches \cite{mitchell1993analogy, gentner1983structure, forbus2017extending, holyoak1989analogical}, each of which instantiated different cognitive theories of analogical reasoning. These theories' standard methodology involves transforming target and base subjects into symbolic representations and identifying conceptually related elements subject to certain constraints. However, these conventional methods require human-crafted rule templates in advance, resulting in the provably NP-Hard computational complexity \cite{veale1997competence} and thereby indicating the inability to generalize new rules. Therefore, several works \cite{hagiwara1991analogical, lu2019seeing,honda2021analogical} have been proposed to employ learning-based methods to enhance inference in analogical reasoning. However, these methods' reasoning still needs to rely on manually designed templates or exploiting database rules. Furthermore, based on the structure-mapping theory (SMT) \cite{gentner1983structure}, a recent work \cite{crouse2020neural} is proposed to abstract the target and base subjects into graph representations and decompose the graph into local relational paths for discovering analogically similar paths in the other graph. However, for the ease of computation, this approach simplifies the whole analogical reasoning problem into individual reasoning problems unaware of each other, which hence degrades the reasoning capability.

\begin{figure*}[!t]
    \centering
    \includegraphics[width=0.9\textwidth]{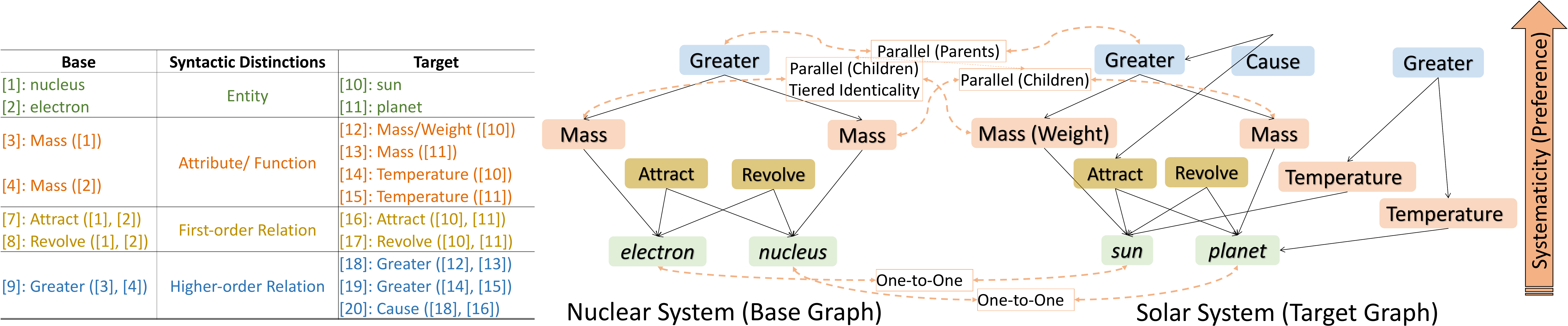}
    \vspace{-1mm}
    \caption{The illustration of analogical reasoning regulated by Structure-Mapping Theory (SMT): two graph representations of the Rutherford model of the atom (left) and the solar system (right).}
    \label{fig: smt_example}
    \vspace{-4mm}
    \end{figure*}

As a high-end human-exclusive intelligence, analogical reasoning is still quite an open area for artificial intelligence, with several critical obstacles in the way of its advancement. \emph{1) Difficulty in locating structural alignment between the base and target subjects.} Analogical reasoning first requires determining a mapping between two relational representations (referred to as the base and target), which is denoted as structural alignment. This mapping must be structurally consistent and include as many element-wise correspondences as possible between two relational representations. Following the setting in SMT by representing both base and target as vertex-labeled Directed Acyclic Graphs (DAGs), analogical reasoning is conceptually similar to discovering a subgraph in the target DAG that is isomorphic to the base DAG. However, identifying isomorphic subgraphs could be problematic since it is a provably \emph{NP-Complete} problem \cite{fortin1996graph}. \emph{2) Difficulty in considering unique constraints throughout the correspondence inference.} Within the analogy derivation framework of SMT, several hard constraints ought to be considered in order to keep the analogical validity of the identified mapping between the base and target DAGs, and these constraints are denoted as relational alignment. For example, SMT regulates each element of the base and target DAGs can only exist at most in one correspondence, and a pair of nodes from the base and target DAG must form a correspondence if their descendant nodes are in correspondence, etc. Without hand-crafted rules or node-wise selection procedures, it is hard to design a learning-based approach to identify suitable correspondence pairs. 

In coping with these challenges, we propose an end-to-end inference framework, namely \underline{Deep} \underline{G}raph learning for \underline{A}nalogical \underline{R}easoning (DeepGAR), to identify the set of correspondences between the base and target subjects in an end-to-end optimization manner. Notably, we employ a directed acyclic graph structure to represent target and base subjects as formulated in SMT. To deal with the first challenge, DeepGAR constructs a node embedding space to incorporate geometric ordering constraints corresponding to subgraph relations to guide the inference objective for effectively identifying the structure alignment. To tackle the second challenge, DeepGAR derives a unified objective that is capable of locating analogical correspondences with constraints to ensure structural and relational alignments. We summarize contributions as follows:
\begin{itemize}[leftmargin=*]
    \item \textbf{We propose a deep graph representation learning-based framework DeepGAR to address the analogical reasoning problem.} DeepGAR leverages deep graph embedding to simplify the combinatorial complexity of searching structural alignments and derives an end-to-end objective function to locate correspondence satisfying all SMT regulations.
    \item \textbf{We develop a new method for constructing a node embedding space for easy structural alignment.} The embedding space can directly induce subgraph relations by comparing the root node embeddings of two DAGs, enabling the efficient identification of structural alignment during the correspondence inference phase.  
    \item \textbf{We conduct extensive experiments on both synthetic and real-world analogical matching datasets.} Compared with existing approaches, DeepGAR achieves superior results in predicting analogical correspondence by outperforming others by on average $15$\% in the F1 score.
\end{itemize}

\section{Problem Formulation}
Structure-Mapping Theory (SMT) \cite{gentner1983structure} provides a graph-theoretic notion that naturally represents both target and base subjects in the form of directed-acyclic graphs (DAG), where each node represents some logical expressions and directed edges carry the hypotactic relations between each expression. We visually demonstrate the SMT in Fig. \ref{fig: smt_example}. Specifically, given a target-directed acyclic graph  $G_T=(V_T, E_T)$ and a base DAG $G_B=(V_B, E_B)$, where the node set size $|V_T| \ge |V_B|$, each node $v_i$ in $V_T$ or $V_B$ is associated with a label $\psi(v_i)$. We use $\psi(v_i)$ to indicate the expression of the node (i.e., an \emph{entity}, an \emph{attribute/function}, or a \emph{relation} as exemplified in Fig.~\ref{fig: smt_example}). In addition, each node is also attached to a unique signature information $\omega(v_i)$ to further clarify the semantic meaning of $v_i$ (e.g., $v_i$ is an \emph{entity} with meaning $\omega(v_i)=``sun"$, or $v_i$ is an \emph{attribute} with meaning $\omega(v_i)=``Mass"$). 

\noindent\textbf{Problem Formulation.} The problem of analogical matching aims to identify a subgraph $G_S=(V_S, E_S), G_S\subseteq G_T$ in the target DAG $G_T$ where each element (i.e., node) $v_S \in V_S$ can form a bijective correspondence to an element $v_B\in V_B$. Specifically, a correspondence should satisfy the pairwise-disjunctive constraint imposed by the \emph{structural alignment} and \emph{relational alignment}. The \emph{structural alignment} requires the subgraph $G_S$ to be isomorphic to $G_B$: $\exists$ bijection $f: V_S\mapsto V_B$ such that $(f(u_B), f(v_B))\in E_B$ if $(u_S, v_S) \in E_S$. The \emph{relational alignment} further regulates the bijection $f$ to satisfy $\phi(f(u_B)) = \phi(u_T), \forall \:u_B \in V_B$ and $\forall \:u_T \in V_T$ as well as the following rules regulated by SMT \cite{gentner1983structure}:
\begin{enumerate}
    \item \emph{Parallel Connectivity}: Two expressions can be in correspondence with each other only if their arguments are also in correspondence with each other.
    \item \emph{One-to-One}: Each element (i.e., node) of the base and target DAG can be a part of at most one correspondence.
    \item \emph{Tiered Identicality}: Relations of expressions in a correspondence must match identically between $G_S$ and $G_B$, but function-typed expressions need not if their correspondence supports parallel connectivity.
    \item \emph{Systematicity}: Preference should be given to mappings with more deeply nested expressions.
\end{enumerate}

\begin{definition}[Analogical Reasoning] In all, the problem can be mathematically stated as follows: Given any base DAG $G_B$ and target DAG $G_T$, we aim at finding an alignment matrix $X\in \{0, 1\}^{|V_B|\times |V_T|}$: $X_{ij}=1$ that matches any node in $G_B$ with a node in $G_T$, following the structural alignment and relational alignment defined above. Here, $X_{ij}=1$ denotes $v_i\in G_B$ matches $v_j\in G_B$; and $X_{ij}=0$, otherwise. \label{df: 1}
\end{definition}

To understand the SMT, we provide an example of SMT-regulated analogical matching between the Rutherford model of the atom and the solar system in Fig. \ref{fig: smt_example}. As illustrated in the figure, a subgraph in the solar system can be located to form an analogical matching to the atom system, where each element (i.e., node) in the atom system can have a correspondence with another node in the solar system (e.g., \texttt{([1]}, \texttt{[10])} and \texttt{([2]}, \texttt{[11])}). The \emph{one-to-one} regulates each element is added up to one correspondence. The \emph{parallel connectivity} restricts correspondence between elements if their parents are in correspondence. For example, \texttt{([9]}, \texttt{[18])} is one correspondence, and their child nodes \texttt{([3]}, \texttt{[12])} and \texttt{([4]}, \texttt{[13])} are also in correspondence. The \emph{Tiered Identicality} further ensures two semantically similar elements can form a correspondence. In this example, mass and weight are similar signatures for \texttt{[12]}, and both labels could allow the correspondence \texttt{([3]}, \texttt{[12])}. Finally, \emph{systematicity} would favor the matched subgraph that has a relatively large node depth since it could result in a larger correspondence set. For example, both \texttt{[18]} and \texttt{[19]} are candidates to form the correspondence with \texttt{[9]}, but the \emph{systematicity} would select \texttt{[18]} since it could derive more relations: e.g., \texttt{[20]}.

However, identifying the analogical correspondences is not a trivial task due to the following challenges. Firstly, searching for structural alignments in $G_T$ that matches $G_B$ is extremely complex. This problem is known as \emph{NP-Complete} \cite{fortin1996graph} and is equivalent to the subgraph isomorphism problem (i.e., finding a bijection $f: V_S\mapsto V_B$). Secondly, SMT has more rigorous criteria (i.e., parallel connectivity, one-to-one, tiered identicality, and systematicity) regulated by SMT for mining analogical correspondences. Thus, designing an objective function to identify the analogical correspondence between $G_T$ and $G_B$ further imposes novel challenges.

\section{Model}
This section focuses on describing our proposed DeepGAR approach to tackle all challenges mentioned above, which 1) constructs an embedding space that directly captures the subgraph relation by node embeddings in latent embedding space, 2) designs a novel inference objective to search for relational alignments that satisfy the constraints of SMT and fulfill both structural and relational alignment requirements.

\subsection{Embedding Space Construction with Geometric Constraint}\label{sec: embedding}
\label{sec:embedding_space}
\noindent\textbf{Overview.} To avoid exhaustively searching for graph isomorphism on $G_T$, we aim to learn graph embedding to handle the original problem in continuous space. The embedding space preserves the geometric ordering and provides guidance in locating structural matching. 

\begin{figure}[!t]
\centerline{\includegraphics[width=0.43\textwidth]{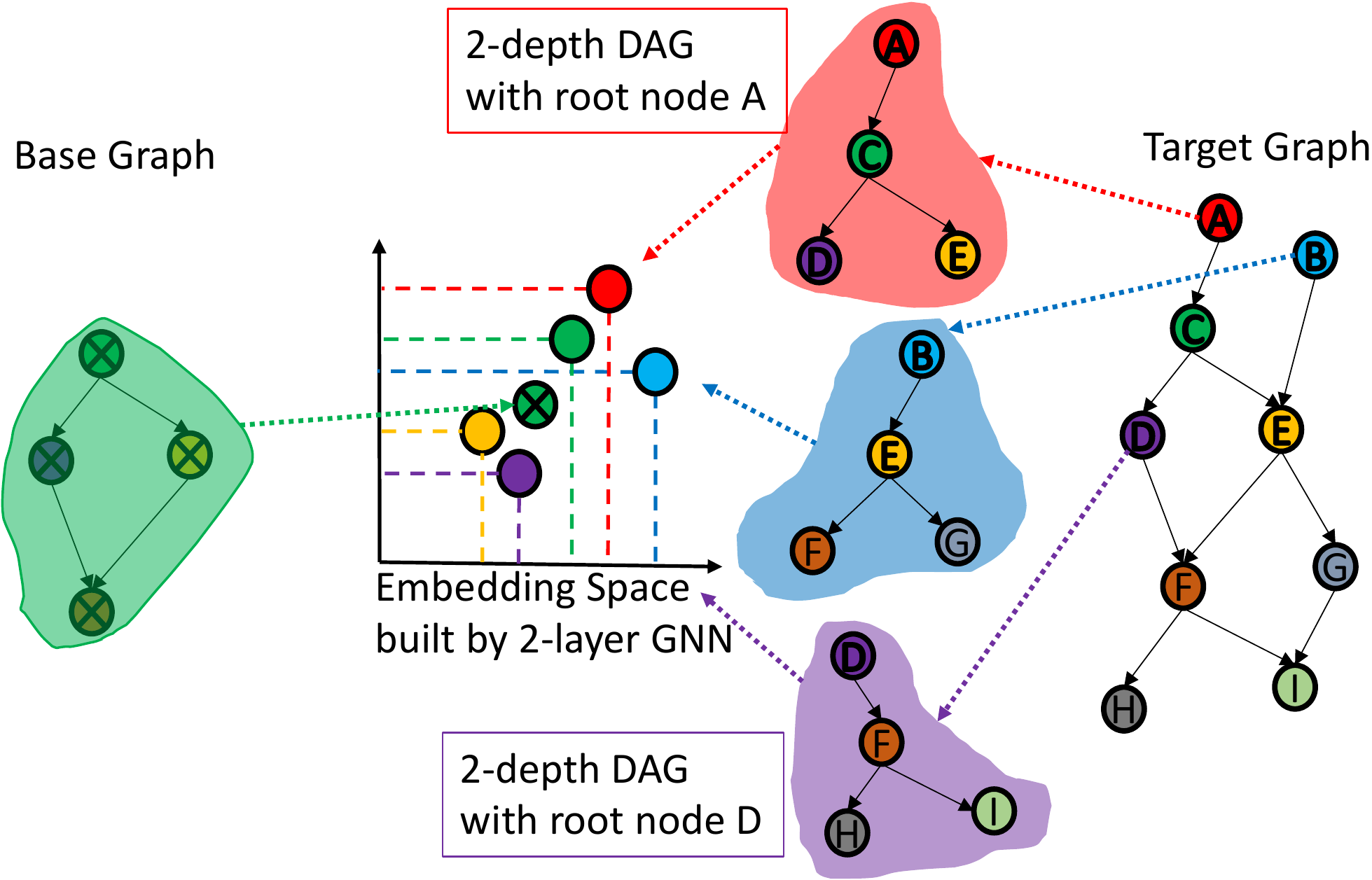}}
\caption{Subgraph-relation between node embeddings in the latent space: the embedding of each node is calculated by a $k$-layer GNN model, and the subgraph relation can be reflected through node embeddings by a designed ordering constraint.}
\label{fig: embedding}
\vspace{-6mm}
\end{figure}

\noindent\textbf{Geometric Ordering Preserved Embedding Space.} We first leverage the GNN model to obtain the embedding of each node. Specifically, we adopt a $K$-layer GNN that aggregates each node information within $K$-hop neighbors through sum pooling. Specifically, the $k$-th layer transformation is:
    \begin{align} \label{eq: gnn}
        h_v^{(k)}=\xi(h_v^{(k-1)}, g(\{h_u^{(k-1)}:u\in \mathcal{N}(v)\})), \: k\in[1, K],
    \end{align}
    where $h_v^{k}$ is the $k$-th layer feature of node $v$, and $h_v^{0} = [\phi(v);\omega(v)]$ is the concatenation of the label $\phi(v)$ and the signature $\omega(v)$ of each node. More specifically, $\phi(v)$ is the one-hot vector describing the node label (e.g., entity or relation) and $\omega(v)$ is the vectorized word embedding obtained by pre-trained language models (e.g., BERT~\cite{DBLP:journals/corr/abs-1810-04805}). Both $\xi(\cdot)$ and $g(\cdot)$ are aggregation function (i.e., sum pooling) to aggregate information from the node $v$ and its $K$-hop neighbors. DeepGAR then constructs an embedding space that could preserve a geometric ordering over subgraphs from node embeddings. Specifically, given a DAG with root node $u$, by using a $K$-layer GNN to embed $u$, we are essentially condensing the $K$-layer neighborhood information of $G_u$ around the root node $u$. Thus, embedding $u$ is equivalent to embedding $G_u$ (a $K$-hop subgraph rooted at node $u$), and by comparing embeddings of two nodes $u$ and $v$, we are essentially comparing the structure of subgraphs $G_u$ and $G_v$. We demonstrate an example in Fig. \ref{fig: embedding}, where the target DAG can be decomposed into many $2$-hop subgraphs with different root nodes. If node $A$ is the parent of node $C$, then the $2$-hop DAG rooted at node $C$ is the subgraph of the $2$-hop DAG rooted at node $A$. We aim to reflect such geometric ordering information directly through node embedding. Specifically, if node $A$ is the parent of node $C$, the embedding $h_C$ is constrained to be on the ``lower-left'' of $h_A$ in the embedding space:
\begin{align}\label{eq: subgraph_relation}
    h_A - h_C \succeq 0, \:\text{if } G_A \subseteq G_C.
\end{align}
Such a well-structured embedding space allows efficiently navigating to the most suitable structural alignments in $G_T$ by directly comparing their root node embeddings without enumerating local DAG structures.


\noindent\textbf{Training Strategy.} During the training phase of constructing the constrained embedding space, we decompose the target DAG $G_T$ as well as the base DAG $G_B$ into many small overlapping DAGs with the depth $K$ ($K$-hop). We denote $\Gamma$ as the set of positive pairs that contains graph pairs $(G_u, G_v)$ has the subgraph relation, and $N$ is denoted as the negative example set. We train the Graph Isomorphism Network (GIN)~\cite{xu2018powerful} that produces the ordering-constrained embeddings $h_u$ and $h_v$ using the max-margin loss:
\begin{align}\label{eq: max-margin}
    \mathcal{L}_e = &\sum_{(h_u, h_v)\in \Gamma}D(h_u, h_v) + \sum_{(h_u, h_v)\in N}\max\{0, \alpha-D(h_u, h_v)\},\nonumber\\
    &\quad\text{s.t. } D(h_u, h_v)=\rVert\max\{0, h_u-h_v\}\rVert^2_2,
\end{align}
where $D(h_u, h_v)$ denotes the magnitude to constrain the subgraph relation violation. For positive pairs, we want to minimize $D(h_u, h_v)$ such that all the elements in the node embedding $h_u$ are less than the corresponding elements in the node embedding $h_v$ if $G_u \subseteq G_v$. For negative pairs, we utilize a regularizer to regulate the amount of violation $D(h_u, h_v)$ to be at least $\alpha$ to avoid zero loss. 

\subsection{Unified Objective for Locating Analogical Matching}
The embedding space constructed in Section~\ref{sec:embedding_space} has enabled us to locate structural alignment by only comparing the root nodes' embedding of two DAGs, which addresses the structural alignment required for the analogical matching problem defined in Definition \ref{df: 1}. In this section, we aim to propose the objective function of analogical matching that searches for the alignment matrix $X$, by further handling the relational alignment that consists of the requirements of \emph{parallel connectivity}, \emph{one-to-one}, \emph{tiered identicality}, and \emph{systematicity}. 

\noindent\textbf{Parallel Connectivity.} The first requirement enforces the correspondence between nodes should exist if their parents are also in correspondence, which implies the structural correspondence between $G_T$ and $G_B$. In other words, if we could locate a subgraph $G_S \subseteq G_T$ that is isomorphic to $G_B$, this requirement can naturally be satisfied. Therefore, we optimize the following objective:
\begin{align*}
    \min\nolimits_{X}\Vert XA_TX^{\intercal}-A_B\Vert^2_F,
\end{align*} where $\rVert \cdot \rVert_F$ stands for the Frobenius norm. We minimize the empirical loss to make sure the identified subgraph $G_S \subseteq G_T$ has the same structure of $G_B$. In addition, by employing the constructed embedding space as guidance, we add a constraint $X\cdot h_{v_T} \succeq h_{v_B}$ on the objective to transform the structural alignment problem as a node-level task by using node embeddings to predict whether the set of nodes $X\cdot h_{v_T}$ selected in $G_T$ can match be matched to the $h_{v_B}$. Note that both $h_{v_B}$ and $h_{v_T}$ are root node embedding of $G_B$ and $G_T$ with $v_B$ and $v_T$ being the root nodes, respectively.


\noindent\textbf{One-to-one.} The second requirement in SMT restricts each element in the alignment matrix to be a member of at most one correspondence. In other words, the alignment matrix $X$ needs to have one and only one matching on each element in $G_B$ (i.e., $\rVert X_{i,*}\rVert_1 = 1$). In this work, we propose to impose the orthogonal constraint (i.e., $X^{\intercal}X=I$) on the alignment matrix, to add the restriction of ensuring there is one and only one matching on each row to meet the \emph{one-to-one} requirement.

\noindent\textbf{Tiered Identicality.} The third requirement suggests that a correspondence between two nodes is also allowed in $G_T$ if their parent node has the semantically similar meaning as the corresponding parent node in $G_B$. Each node is attached to a signature and a label, and we leverage word embedding to obtain latent representation for each node. If two nodes are semantically similar, their latent representations would also be close. The GNN model would naturally consider projecting semantically similar nodes closely in the constructed embedding space. Therefore, this constraint can also be fulfilled by the constraint $X\cdot h_{v_T} \succeq h_{v_B}$ during the optimization.


\noindent\textbf{Systematicity.} The last property of SMT favors larger correspondence sets over smaller ones in order to look for larger and deeper matches. Therefore, if there are multiple valid subgraphs on the target DAG $G_T$, SMT would prefer the subgraph that has the relatively higher average node depth. Therefore, the regularization term $-\rVert X\cdot d_T\rVert^2_2$ could be added to make the alignment selection be aware of the node depth. Note that $d_T \in \mathbb{R}^{|V_T|}_{\ge 0}$ is the vector of node depth in $G_T$.

Therefore, we derive the following objective function $\mathcal{L}$ as well as its augmented Lagrangian form that relaxes the hard constraints in the objective. 
\begin{align}\label{eq: inference_obj_full}
    \mathcal{L}&=\min\nolimits_{X}\Vert XA_TX^{\intercal}-A_B\Vert^2_F-\rVert X\cdot d_T\rVert^2_2\\
    &\quad \quad \text{s.t. } \: X\cdot h_{v_T} \succeq h_{v_B}, \: X^{\intercal}X=I,\nonumber\\
    &=\min\nolimits_{X}\Vert XA_TX^{\intercal}-A_B\Vert^2_2+\lambda_1\rVert J\cdot d_T - X\cdot d_T\rVert_2 \nonumber\\
    &+\lambda_2\Vert\max(0,h_{v_B}- X\cdot h_{v_T}))\Vert^2_2 + \lambda_3\rVert I-X^{\intercal}X\rVert_2,\nonumber
\end{align}
where we minimize $\Vert XA_TX^{\intercal}-A_B\Vert^2_F$ and additional constraints to ensure the selection satisfies the relational alignment regulated by the SMT. Furthermore, $\lambda_1$, $\lambda_2$, and $\lambda_3$ are non-negative regularization hyperparameters. When the optimization is done, we apply the sigmoid function to allow the value in each cell $X_{i,j}\in[0, 1]$ to represent the node selection probability. We then output the most likely nodes to form the final correspondence matrix $X$.



\newcolumntype{g}{>{\columncolor{Gray}}c}
\begin{table*}[t]
\centering
\resizebox{\textwidth}{!}{%
\begin{tabular}{@{}c|ccccc|ccccc|ccccc|ccccc@{}}
\toprule
        & \multicolumn{5}{c|}{Synthetic}         & \multicolumn{5}{c|}{Oddity}      & \multicolumn{5}{c|}{Moral}   & \multicolumn{5}{c}{Geometric} \\ \midrule
Methods & ACC  & RE     & PR     & F1     & AUC    & ACC  & RE     & PR     & F1     & AUC    & ACC  & RE     & PR     & F1     & AUC    & ACC  & RE     & PR     & F1     & AUC    \\ \midrule
NeuroMatch     & 0.903 & 0.346 & 0.533 & 0.419 & 0.803 & 0.914 & 0.433 & 0.502 & 0.465 & 0.685 & 0.902 & 0.513 & 0.548 & 0.529 & 0.765 & 0.896 & 0.334 & 0.462 & 0.388 & 0.701  \\\midrule
AMN     & 0.923 & 0.281 & \textbf{0.763} & 0.411 & 0.844 & 0.917 & 0.231 & \textbf{0.632} & 0.338 & 0.725 & \textbf{0.919} & 0.335 & \textbf{0.746} & 0.462 & 0.813 & 0.897 & 0.316 & \textbf{0.694} & 0.434 & \textbf{0.774}  \\\midrule
\rowcolor{Gray}
DeepGAR & \textbf{0.931} & \textbf{0.623} & 0.585 & \textbf{0.603} & \textbf{0.846} & \textbf{0.937} & \textbf{0.454} & 0.497 & \textbf{0.475} & \textbf{0.729} & 0.916 & \textbf{0.577} & 0.609 & \textbf{0.593} & \textbf{0.821}  & \textbf{0.933} & \textbf{0.429} & 0.474 & \textbf{0.450}  & 0.747  \\
\bottomrule
\end{tabular}
}
\caption{Performance of correspondence prediction over comparison methods. (Best is highlighted with bold.)}
\vspace{-3mm}
\label{tab: evaluation_1}
\end{table*}

\newcolumntype{g}{>{\columncolor{Gray}}c}
\begin{table*}[t]
\centering
\resizebox{\textwidth}{!}{%
\begin{tabular}{@{}c|ccccc|ccccc|ccccc|ccccc@{}}
\toprule
        & \multicolumn{5}{c|}{Synthetic}         & \multicolumn{5}{c|}{Oddity}      & \multicolumn{5}{c|}{Moral}   & \multicolumn{5}{c}{Geometric} \\ \midrule
Methods & ACC  & RE     & PR     & F1     & AUC    & ACC  & RE     & PR     & F1     & AUC    & ACC  & RE     & PR     & F1     & AUC    & ACC  & RE     & PR     & F1     & AUC    \\ \midrule
AMN     & 0.845 & 0.902 & 0.843 & 0.872 & 0.405 & 0.932 & \textbf{0.933} & 0.904 & 0.918 & \textbf{0.514} & 0.862 & 0.764 & \textbf{0.913} & 0.832 & 0.461 & \textbf{0.936} & \textbf{0.911}  & \textbf{0.884} & \textbf{0.897} & 0.497  \\\midrule \rowcolor{Gray}
DeepGAR & \textbf{0.851} & \textbf{0.913} & \textbf{0.866} & \textbf{0.889} & \textbf{0.432} & \textbf{0.936} & 0.926 & \textbf{0.910}  & \textbf{0.919} & 0.508 & \textbf{0.884} & \textbf{0.812} & 0.896 & \textbf{0.852} & \textbf{0.493} & 0.892 & 0.903 & 0.865 & 0.884 & \textbf{0.502}  \\
\bottomrule
\end{tabular}
}
\caption{Performance of candidate inference over comparison methods. (Best is highlighted with bold.)}
\vspace{-5mm}
\label{tab: evaluation_2}
\end{table*}

\section{Experiment}
To investigate the effectiveness of DeepGAR, we compare its performance with existing approaches in both synthetic and real-world datasets.

\subsection{Experiment Setup}
\noindent\textbf{Comparison Methods.} We compare the performance of DeepGAR on various experiments against two sets of methods. 
\begin{itemize}[leftmargin=*]
    \item \textit{Analogical Matching Approaches.} \emph{AMN} \cite{crouse2020neural} is the first and only method that applies deep learning in identifying analogical correspondences that are consistent with the principles of SMT. It learns the node embedding by LSTM-based encoder, and use node embeddings to iteratively search for correspondence and available candidates.
    \item \textit{Subgraph Matching Approaches.} \emph{NeuroMatch} \cite{lou2020neural} is the first and STOA work applying deep learning and graph neural network to impose a geometric ordering between node embeddings to attain efficiency in solving subgraph isomorphism, which is conceptually similar to analogical reasoning in terms of structural alignment.
    \end{itemize}

\noindent\textbf{Data.} We demonstrate the performance of our proposed DeepGAR over both synthetic data and real-world data. \emph{1) Synthetic.} The synthetic analogical examples are constructed based on DAG generation, and we follow the procedure described in \cite{crouse2020neural} to generate synthetic DAGs with depth $k$ sampled from $[2, 7]$. Analogical pairs are sampled from each generated DAG. \emph{2) Visual Oddity.} There are $3,405$ analogical comparisons to explore cultural differences in geometric reasoning in \cite{dehaene2006core}. \emph{3) Moral Decision Making.} This dataset \cite{dehghani2008moraldm} aims to understand moral decision-making and compare previously solved cases to novel situations with $420$ analogical comparisons. \emph{4) Geometric Analogies.} The object of this dataset is to select a geometric figure that best completes the analogy from an encoded set of possible answers. We follow the processing steps in \cite{forbus2011cogsketch, crouse2020neural} and extract $866$ pairs. In order to test each model's generalization capability and its usage in real-world analogical reasoning tasks, we only train each comparison method on the synthetic dataset. The trained models are then deployed to make predictions on each dataset without fine-tuning. Due to the limited space, more detailed explanations of each dataset can be found along with the code.

    \noindent\textbf{Implementation Details and Evaluation.} DeepGAR is flexible in terms of the GNN model used for the embedding step, and we particularly choose $K$-layer GIN \cite{xu2018powerful} model with skip layers to encode DAG information. We choose Adam with learning rate $0.001$ for optimizing the Equation \eqref{eq: inference_obj_full}, and hyper-parameters are set to be $\lambda_1= 1e-3$,  $\lambda_2= 1e-1$,  $\lambda_3= 1e-3$. We uniformly sample $50\%$ of the graphs in the synthetic dataset to train the GIN model. The analogical matching problem is a classification task to distinguish whether the identified subgraph $G_S \in G_T$ is the correct analogical matching to $G_B$. We thus adopt Accuracy (ACC), Recall (RE), Precision (PR), F1-Score (F1), and ROC-AUC (AUC) to evaluate whether the identified subgraph is the correct matching to the base.    
    

\subsection{Correspondence Prediction}
We first demonstrate the performance of the correspondence prediction among all approaches, and the results are described in Table \ref{tab: evaluation_1}. Note that the data generator generates ground truth analogical correspondences when evaluating the synthetic data. For all real-world datasets, the comparison set of correspondences is computed by a rule-based algorithm: structure mapping engine (SME) \cite{falkenhainer1989structure}, which is built on SMT and is the most widely accepted computational model of SMT.

As shown in the table, DeepGAR performs the best among all comparison methods except for the precision score, and the F-$1$ excels other methods by, on average $15$\%. We draw a couple of interesting observations. Firstly, the subgraph matching algorithm - NeuroMatch performs the worst among all comparison methods since it is not designed for the analogical reasoning task. During the inference phase, NeuroMatch can only identify structural matching between the base DAG and target DAG. Without considering all the constraints regulated by the SMT, NeuroMatch cannot handle scenarios: 1) there exist multiple valid structural matchings but only one satisfies the SMT; 2) there is only a part of the base DAG can be matched to the target DAG (i.e., partial matching). In addition, though AMN and DeepGAR have similar performance in ACC and AUC, the imbalance between the PR and RE further demonstrates that AMN tends to generate less but accurate correspondences, which results in the imbalanced PR and RE. Moreover, the AMN-generated correspondences are not organized into a graph structure, which prevents us from learning the identified analogy's latent and potentially hierarchical structure.

\subsection{Candidate Inference Prediction}
Candidate inferences are statements from the base projected into the target to fill in missing structure \cite{crouse2020neural, holyoak1989analogical}. Given a set of correspondences, candidate inferences are constructed from statements in the base supported by expressions in the correspondence set but are not included in the correspondence set. In many real-world analogical reasoning problems, every non-correspondence node can become a candidate inference (which can lead to inflated precision and recall values). Thus, we utilize the correspondences that are accurate but not formed in a graph structure as the ground truth and report the classification performance. Due to the designed pairwise node selection process, AMN could naturally output candidate correspondence. In DeepGAR, after we make the correspondence selection, we output the most likely correspondence as candidates. Since NeuroMatch is not designed for handling such a task, we only compare the performance between DeepGAR and AMN, and report results in Table \ref{tab: evaluation_2}. As exhibited in the table, DeepGAR is still effective in mining candidates. Due to its design, AMN has advantages in determining the candidates since it tends to produce fewer correspondences. Compared to AMN, DeepGAR provides consistently competitive and, most of the time, better performance across all datasets.

\begin{table}[]
\centering
\resizebox{0.47\textwidth}{!}{%
\begin{tabular}{@{}c|c|cccc@{}}
\toprule
\multirow{2}{*}{} & \multicolumn{1}{c|}{\multirow{2}{*}{\begin{tabular}[c]{@{}c@{}}Training\\ Runtme\end{tabular}}} & \multicolumn{4}{c}{Inference Runtime}  \\ \cmidrule(l){3-6} 
                  & \multicolumn{1}{c|}{}                                                                           & Synthetic & Oddity & Moral & Geometric \\ \midrule
NeuroMatch        &          $2547.62$  &                                                       ${3414.71}$       &   $5618.22$  &  $1336.23$ &   $2671.56$          \\
AMN               &          $18655.31$                                                                                       &  $54142.31$  & $121743.62$  & $22354.47$  & $34142.17$  \\ \rowcolor{Gray}
DeepGAR            &        ${1435.71}$                                                                                         &  $20344.12$ & $43462.51$ &  $8679.41$    &   $15634.22$    \\ \bottomrule
\end{tabular}%
}
\caption{We separately record the training runtime and inference runtime on each dataset.}
\vspace{-3mm}
\label{tab: runtime}
\end{table}

\begin{figure}[!t]
\vspace{-3mm}
		\subfloat[AMN - Moral]{\label{fig: geo_amn}
			\hspace{-3mm}\includegraphics[width=0.16\textwidth]{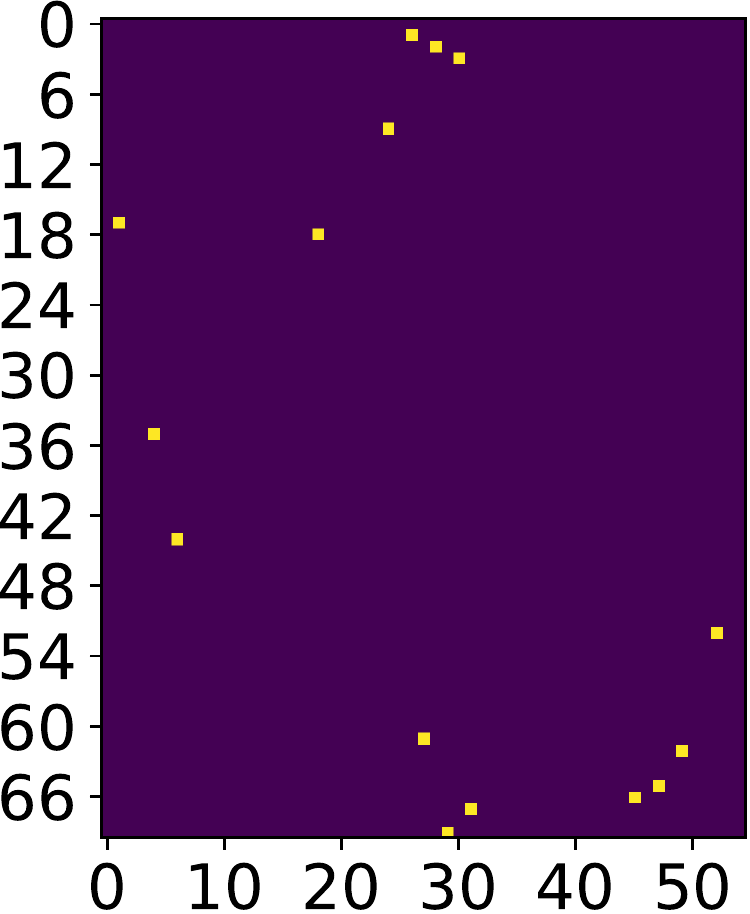}}
		\subfloat[DeepGAR - Moral]{\label{fig: geo_DeepGAR}
			\includegraphics[width=0.16\textwidth]{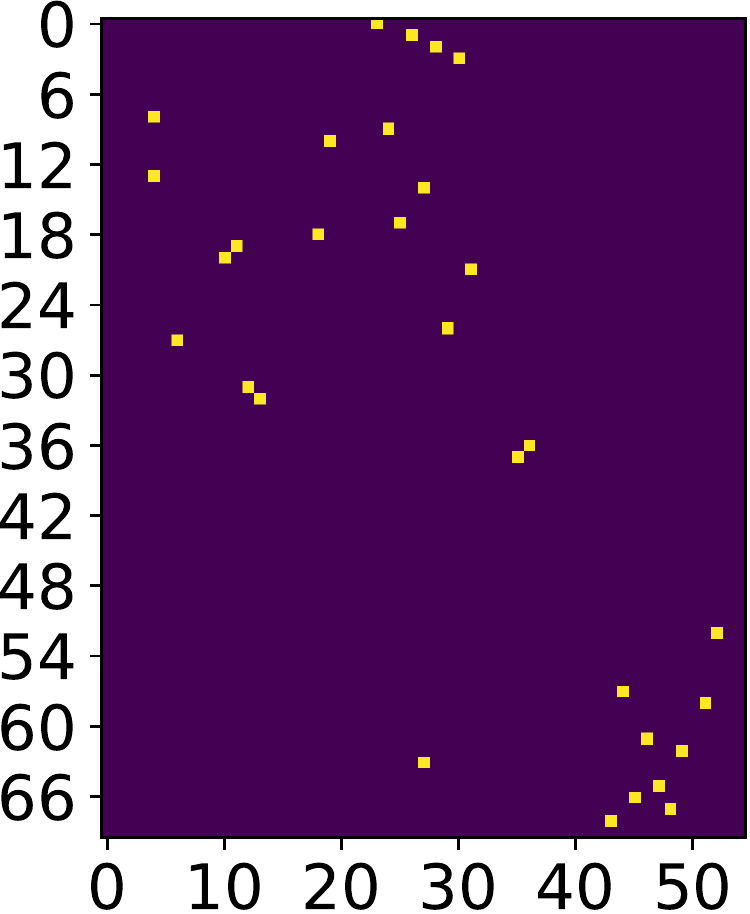}}
		\subfloat[True - Moral]{\label{fig: geo_true}
			\includegraphics[width=0.16\textwidth]{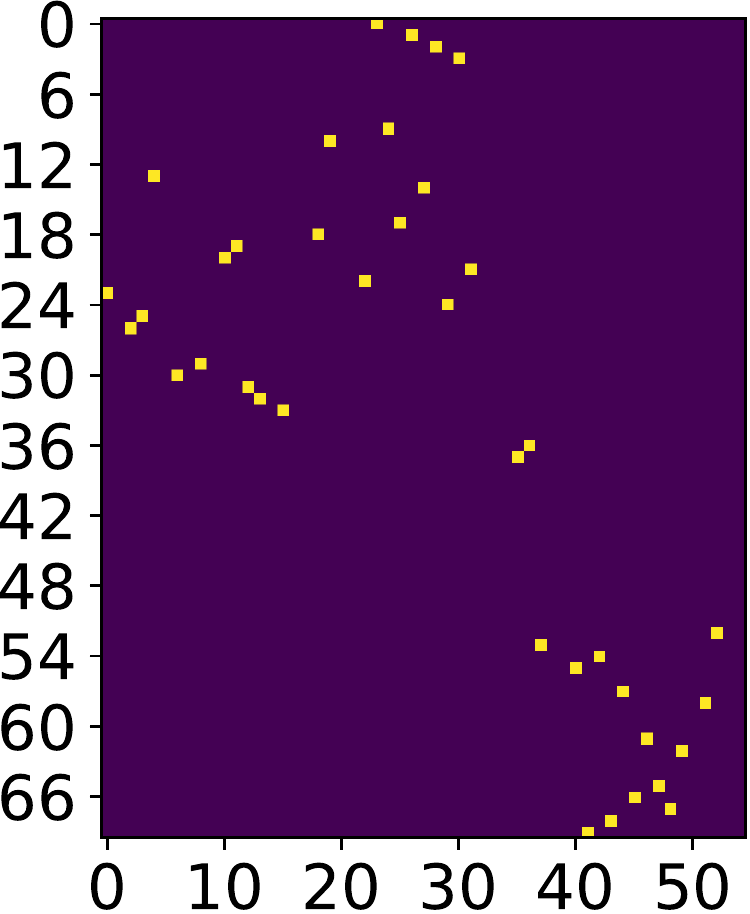}}
		\vspace{-2mm}
		\caption{The comparison of generated analogical alignment matrix in the Moral dataset.}
		\vspace{-7mm}
		\label{fig: viz}
	\end{figure}

\subsection{Scalability Analysis}
To analyze the scalability of the proposed model, we record the average runtime for the training of all comparison methods. We repeat the experiments $10$ times until convergence and report the average runtime of training and inference on each dataset in Table \ref{tab: runtime}, respectively.

As depicted in Table \ref{tab: runtime}, AMN takes more than $7$ times of the runtime against DeepGAR and NeuroMatch in training, which is mainly due to its Transformer-based neural network and LSTM-based graph encoder. DeepGAR has the overall lowest training time and excels over other methods by an evident margin. Furthermore, DeepGAR only needs to construct a geometric ordering constrained embedding space, which saves a significant amount of time and computational resources compared to other methods. In terms of the inference runtime, NeuroMatch has the generally fastest runtime since it does not involve any optimization steps. AMN still exhibits the longest inference runtime since it leverages the Transformer-based node-wise correspondence predictor, which requires a lot more reasoning time due to the need to enumerate the graph structure of both $G_T$ and $G_B$.

\subsection{Visualization}
In addition to our main results, we provide qualitative examples of DeepGAR’s outputs on real analogy problems. We randomly choose one analogical matching case from the Moral dataset (i.e., the largest dataset) and visualize the identified analogical matching matrices. We illustrate the comparison between DeepGAR, AMN, and the ground truth in Fig. \ref{fig: viz}. The results in the visualization align with the analysis we draw on the correspondence prediction. The alignment matrix generated by AMN contains isolated correspondences, while DeepGAR consistently generates structurally similar alignment matrices to real ones across different real-world scenarios.

\section{Conclusion}
In this paper, we propose an end-to-end learning framework - DeepGAR, to produce analogical correspondences consistent with constraints regulated by SMT. Specifically, DeepGAR firstly uses graph neural networks and geometric embeddings to learn subgraph relationships. We then derive a unified objective that can locate analogically similar representation in the target with the guidance of constraints regulated by SMT and the constructed embedding space. Extensive experiments and analyzes are conducted to demonstrate the strength of DeepGAR in various synthetic and real-world scenarios.

\section*{Acknowledgement}
    This work was supported by the National Science Foundation (NSF) Grant No. 1755850, No. 1841520, No. 2007716, No. 2007976, No. 1942594, No. 1907805, a Jeffress Memorial Trust Award, Amazon Research Award, NVIDIA GPU Grant, and Design Knowledge Company (subcontract number: 10827.002.120.04).

\bibliographystyle{IEEEtran}
\bibliography{reference}

\end{document}